\documentclass[lettersize,journal,twoside]{IEEEtran}
\usepackage{amsmath,amsfonts}
\usepackage{algorithmic}
\usepackage{algorithm}
\usepackage{array}
\usepackage[caption=false,font=normalsize,labelfont=sf,textfont=sf]{subfig}
\usepackage{textcomp}
\usepackage{stfloats}
\usepackage{url}
\usepackage{verbatim}
\usepackage{graphicx}
\usepackage{cite}
\usepackage{xcolor,color}
\usepackage{booktabs}
\usepackage{multirow}

\newcommand{\zl}[1]{\textcolor{black}{#1}}
\usepackage{ulem}
\usepackage[colorlinks,linkcolor=black,anchorcolor=black,citecolor=black]{hyperref}
\begin{document}

\title{Lightweight Model Pre-training via Language Guided Knowledge Distillation}

\author{Mingsheng Li*, Lin Zhang*, Mingzhen Zhu*, Zilong Huang, Gang Yu, 

Jiayuan Fan,~\IEEEmembership{Member,~IEEE}, and Tao Chen$^{\dagger}$,~\IEEEmembership{Senior Member,~IEEE}

% IEEE Publication Technology,~\IEEEmembership{Staff,~IEEE,}
        % <-this % stops a space
\thanks{Mingsheng Li, Lin Zhang, Mingzhen Zhu, and Tao Chen are with the School of Information Science and Technology, Fudan University, Shanghai 200433, China. E-mail:  limc22@m.fudan.edu.cn; zhangl22@m.fudan.edu.cn; mzhenz@foxmail.com; eetchen@fudan.edu.cn. 

Zilong Huang and Gang Yu are with the Tencent GY-Laboratory, Shanghai 200000, China. E-mail: zilong.huang2020@gmail.com; iskicy@outlook.com.

Jiayuan Fan is with the Academy for Engineering and Technology, Fudan
University, Shanghai 200433, China. E-mail: jyfan@fudan.edu.cn.

$^{\dagger}$Corresponding author.

*Equal Contribution.
}}

% \markboth{IEEE TRANSACTIONS ON MULTIMEDIA, 2024}%
% {Li \MakeLowercase{\textit{et al.}}: Lightweight Model Pre-training via Language Guided Knowledge Distillation}

\maketitle

%%%%%%%%% ABSTRACT
\begin{abstract}
   This paper studies the problem of pre-training for small models, which is essential for many mobile devices. 
   Current state-of-the-art methods on this problem transfer the representational knowledge of a large network (as a Teacher) into a smaller model (as a Student) using self-supervised distillation, improving the performance of the small model on downstream tasks.
   However, existing approaches are insufficient in extracting the crucial knowledge that is useful for discerning categories in downstream tasks during the distillation process. 
   In this paper, for the first time, we introduce language guidance to the distillation process and propose a new method named Language-Guided Distillation (LGD) system, which uses category names of the target downstream task to help refine the knowledge transferred between the teacher and student. 
   To this end, we utilize a pre-trained text encoder to extract semantic embeddings from language and construct a textual semantic space called Textual Semantics Bank (TSB). Furthermore, we design a Language-Guided Knowledge Aggregation (LGKA) module to construct the visual semantic space, also named Visual Semantics Bank (VSB). 
   The task-related knowledge is transferred by driving a student encoder to mimic the similarity score distribution inferred by a teacher over TSB and VSB. Compared with other small models obtained by either ImageNet pre-training or self-supervised distillation, experiment results show that the distilled lightweight model using the proposed LGD method presents state-of-the-art performance and is validated on various downstream tasks, including classification, detection, and segmentation. We have made the code available at \href{https://github.com/mZhenz/LGD}{https://github.com/mZhenz/LGD}.
\end{abstract}

\begin{IEEEkeywords}
Lightweight model pre-training, language-guided distillation, textual semantics bank, visual semantics banks
\end{IEEEkeywords}

%%%%%%%%% BODY TEXT
\section{Introduction}
\label{sec:intro}

\IEEEPARstart{R}{ecently}, the study of pre-training lightweight (small) models with both a small number of parameters and fast inference speed receives increasing attention~\cite{he2022knowledge, shi2022efficacy, wu2022self}. Among existing small model pre-training methods, the self-supervised distillation (SSD)~\cite{fang2021seed, xu2021bag, liu2022improving} which improves the pre-training performance of small models with a distillation signal from a pre-trained large model, has appeared as a promising solution to this problem, as this pipeline can save the overhead of image labeling and meanwhile maintain reasonable performance. Hence we follow this pipeline to design a small model pre-training approach in this work.

\begin{figure}[t]
\centering
\includegraphics[width=\columnwidth]{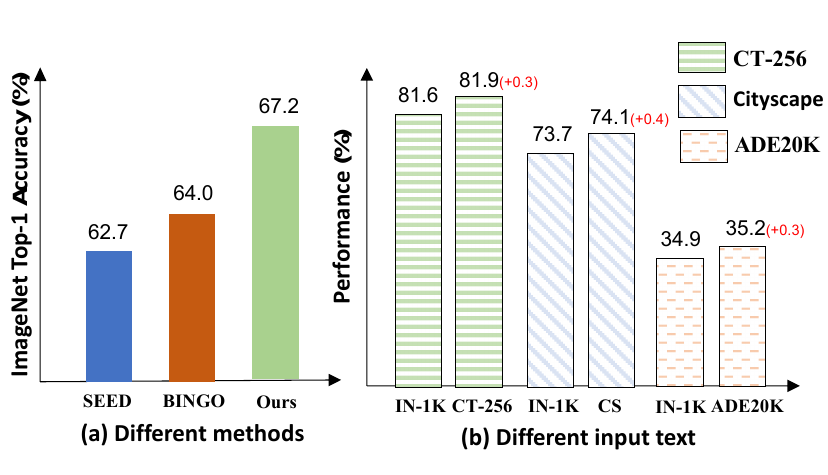}
%\vspace{-10mm}
% \vspace{-5mm}
\caption{(a) Comparison of previous self-supervised distillation (SSD) and the proposed Language-Guided Distillation (LGD) on Imagenet-1k, see more details in sec.~\ref{exp-classification}. (b) Comparison of using different input texts for LGD on ImageNet-1k, then fine-tuning on downstream tasks, e.g., classification on Caltech-256, and segmentation on Cityscapes and ADE20K. Using task-related category names as the input language for knowledge distillation could bring constant improvements, see more details in sec.~\ref{exp-downstream}.}
\label{fig_intro}
\vspace{-3mm}
\end{figure}

Considering the smaller capacity, it is difficult for the student model to fully mimic the outputs of the large teacher model. In other words, the student model may not learn all knowledge of the teacher model. 
An ideal solution is to let the student learn partial but essential knowledge of the teacher, such as the knowledge that is most useful for distinguishing categories in the target tasks. 
Unfortunately, existing SSD methods such as SEED~\cite{fang2021seed} and CompRess~\cite{abbasi2020compress} do not consider this.
Inspired by the success of Contrastive Language-Image Pre-training (CLIP)~\cite{radford2021learning}, which reflects the value of natural language for enhancing the visual semantic features, we propose to distill the essential knowledge of the teacher with the help of language. 
The reason is that the input language could be built on the information of the target task, e.g., the category names in the classification, segmentation, or detection task. Thus, the textual semantic space is also relevant to the target task and can naturally help to identify the most relevant knowledge in the teacher for distillation. Specifically, we make use of the language with the pre-trained text encoder to extract semantic embeddings and build a textual semantic space, also named Textual Semantics Bank (TSB). Utilizing the proposed TSB to enhance distillation and align the outputs of the student and teacher in the target-related textual semantic space can help the lightweight student learn more knowledge of the teacher.

However, the textual semantic feature extracted from the pre-trained text encoder may not be consistent with the corresponding visual feature extracted by the teacher model, which could result in performance degradation in the distillation process. To alleviate this issue, we design a Language-Guided Knowledge Aggregation (LGKA) module to build visual semantic space, also named Visual Semantics Bank (VSB). The VSB shares the same shape as TSB. During the training process, the LGKA module takes a visual feature extracted from the teacher and the TSB as inputs. Then the visual feature is used to momentum update the corresponding feature in the VSB. The correspondence is determined by the similarity between the input visual feature and the TSB. Thus, the VSB is closer to the real distribution of visual semantics than the TSB. The outputs of the student and teacher will be aligned in both the visual and textual semantic spaces. 

Therefore, we design two language-guided loss functions. The first is based on the output consistency constraint of the teacher and student in the textual semantic space, and the second is based on consistency loss after mapping the outputs of the teacher and student to the visual semantic space. It is worth noting that the TSB is fixed, while the VSB is continually updated during the training process.

To this end, we propose a Language-Guided Distillation (LGD) framework, which pre-trains a small model by integrating language guidance into the distillation process. 
Despite the language guidance used, 
we do not use labels, which is the same setting as SSD, but only use unlabeled images and self-determined texts as source data.
As shown in Fig.~\ref{fig_intro}, distilling the same teacher model, the proposed LGD could achieve much better results on the ImageNet-1K dataset with the help of input language. Besides, we could take the corresponding category names according to the target downstream task as language guidance for distilling processing, rather than the fixed one, e.g. category names of ImageNet-1K. It also could bring constant improvements to downstream tasks.

In summary, the main contributions of this paper are summarized below:
\begin{itemize}
\item We propose a novel Language-Guided Distillation (LGD) framework, which is the first attempt to introduce language guidance into the distillation process to pre-train the small model.
\item A Language-Guided Knowledge Aggregation (LGKA) module is developed, which uses language to guide teachers on how to structure the knowledge they pass to students and constrain the learning scope to improve the learning effect. Meanwhile, two losses are designed to maintain the consistency of both the image-to-image and image-to-text relationships between teacher and student features.
\item Thorough experiments are conducted on six datasets and six downstream tasks, which shows the small model pre-trained by the proposed method has better transferability.
\end{itemize}

The remainder of this paper is organized as follows. Section \ref{sec:formatting} reviews the related works on small model pre-training, vision-language model pre-training, and knowledge distillation respectively. In Section \ref{3_method}, we present an overview of the whole framework, including language-guided knowledge aggregation, language-aware knowledge transfer, and optimization objectives. In Section \ref{4_exp}, extensive experiments and analysis are presented to validate the effectiveness of the proposed LGD. Finally, conclusions are drawn in Section \ref{5_con}.

%------------------------------------------------------------------------
%%%%%%%% Related Work
%------------------------------------------------------------------------
%------------------------------------------------------------------------
\section{Related Work}
\label{sec:formatting}

\subsection{Small Model Pre-training}
As a conventional model pre-training way, fully-supervised pre-training~\cite{he2019rethinking} trains a classification network to predict a fixed set of predetermined object categories on a large-scale annotated dataset like ImageNet~\cite{deng2009imagenet}. To alleviate the massive overhead of annotation, self-supervised learning (SSL)~\cite{chen2020simple, he2020momentum, chen2021exploring, self_tip1/wang2023self, self_tip2/liu2022tcgl, self_tip3/ge2023learning, self_tmm1/tao2022self, 
self_tmm2/liu2022self, self_tmm3/zhang2023beyond} aims to dig out good feature representation through the relationship between data samples without labels. However, because of the significant performance drop when the model size decreases,  SSD~\cite{fang2021seed, gao2022disco, xu2021bag, he2022knowledge} is proposed to improve the SSL performance on a small model. Specifically, SSD transfers the learned feature representation from an off-the-shelf pre-trained large model to the student. However, previous SSD methods do not consider the gap in feature representation ability between teacher and student, and they directly transfer all knowledge from the teacher to the student. In this paper, we introduce language guidance to distillation process and use language to constrain the learning scope of the small model.

\subsection{Vision-Language Learning} 
During the past few years, vision-language learning has attracted growing attention~\cite{radford2021learning, jia2021align, m3dbench/li2023m3dbench,ll3da/chen2024ll3da, pre_tmm1/yang2023effective, pre_tmm2/xing2023dual,vote2cap/chen2024vote2cap}. 
As a milestone, CLIP~\cite{radford2021learning} learns high-quality image representations by a simple image text pairing task on a dataset of 400 million (image, text) pairs collected from the internet. 
Motivated by CLIP, several works have emerged to improve the training strategy~\cite{li2022declip, yao2021filip, li2022grounded} or apply the CLIP method to other domains~\cite{wang2021actionclip, rao2022denseclip}. 
However, previous methods directly adopt the CLIP pre-trained large model to downstream tasks, which is unsuitable for practical applications due to the significant computation overhead.
In this paper, for the first time, we transfer the visual and textual representations learned by the CLIP pre-trained large model to the small model through distillation, which reveals a new application of the CLIP pre-trained model. 

\subsection{Knowledge Distillation}
Knowledge distillation usually transfers knowledge from a large teacher to a small student. Previous methods mainly fall into three streams: response-based~\cite{bucilua2006model, hinton2015distilling, zhao2022decoupled}, feature-based~\cite{romero2014fitnets, zagoruyko2016paying, chen2021cross}, and relation-based ~\cite{yim2017gift, tian2019contrastive}. 
Previous works~\cite{fang2021seed, gao2022disco, xu2021bag} have demonstrated that relation-based SSD outperforms response-based and feature-based distillation strategies for self-supervised contrastive pre-trained teacher models, which targets modeling the relationship between features of different samples. 
Similar to SSL works, CLIP also adopts a contrastive learning strategy. But the difference is that CLIP not only models the relationship between image and image but also between image and language (text), which previous works have not considered.
In this paper, we first adopt language guidance to knowledge distillation, which effectively transfers the relation knowledge of image-to-image and image-to-text with language guidance.

\textbf{SSD v.s. LGD.} Here we make a comparison of the self-supervised distillation with  the proposed language-guided distillation. \textbf{Similarly}, they all do not use any label information for distillation. They all support the self-supervised pre-trained model (MoCo~\cite{he2020momentum}), weakly-supervised pre-trained model (CLIP~\cite{radford2021learning}), and fully supervised pre-trained model as the teacher model. ~\textbf{Differently}, LGD takes the self-determined category names (always using the category names from the downstream task) as language guidance, it does not need extra manual labor to annotate. Thus, we think it is fair to compare LGD with other SSD methods. Besides, to evaluate the generalization \zl{ability} of the proposed method, we evaluate our method with the weakly-supervised pre-trained text model (CLIP text encoder~\cite{radford2021learning}) and self-supervised pre-trained text model (BERT~\cite{jacob2019bert}).

%-------------------------------------------------------------------------
%%%%%%%%%%%% METHOD
%-------------------------------------------------------------------------
\section{The Proposed Method}
\label{3_method}

\begin{figure*}[t]
\centering
\includegraphics[width=\textwidth]{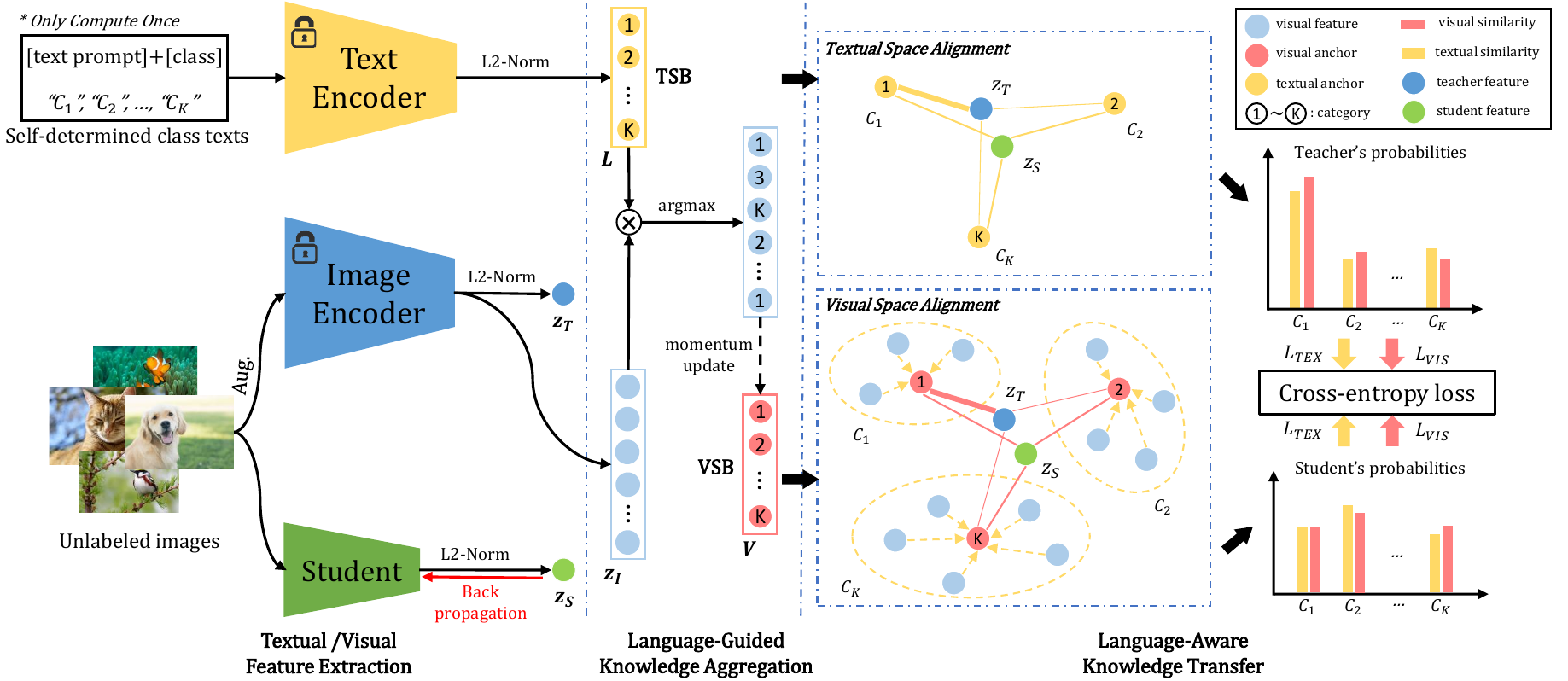} 
\caption{The overview of the proposed Language-Guided Distillation (LGD) framework. First, all the textual features are extracted by feeding \textit{[text prompt] + [class]} into a pre-trained text encoder and stored in the Textual Semantics Bank (TSB) at one time. Besides, visual features are extracted by feeding unlabeled images into a pre-trained image encoder (teacher) and the student. 
Then, a Language-Guided Knowledge Aggregation (LGKA) module is developed to classify the visual features of each batch by the textual anchors, and maintain a Visual Semantics Bank (VSB) to store centroid features of different categories by momentum updating.
At last, to align the feature similarity between the teacher/student feature and anchor in both textual and visual space, the cross-entropy loss is adopted to constrain the visual and textual similarity distribution between teacher and student.
Best viewed in color.
}
%\vspace{-3mm}
\label{fig_method}
%\vspace{-2mm}
\end{figure*}

\subsection{Overview}

The overall framework of the proposed LGD is shown in Fig.~\ref{fig_method}, which consists of two major modules: Language-Guided Knowledge Aggregation (LGKA) and Language-aware Knowledge Transfer. 

Firstly, textual semantic embeddings are extracted by feeding \textit{[text prompt] + [class]} (e.g. "a photo of cat") into the pre-trained text encoder and stored in the Textual Semantics Bank (TSB), denoted as $L$. Here, the semantic embeddings can be regarded as the clustering centers of the self-determined categories. Following CLIP~\cite{radford2021learning}, the prompt engineering and feature ensembling are adopted here so that there is only one textual semantic embedding for each semantic category. Note that the \textit{[class]} can be determined by the knowledge required on downstream task, and textual semantic embeddings only need to be extracted once. Similar to previous SSD works~\cite{fang2021seed, liu2022improving}, unlabeled images after data augmentation are fed into the image encoder (teacher) and student network. 

As mentioned earlier, the textual semantic embeddings extracted from the pre-trained text encoder may not be consistent with the corresponding visual feature extracted by the teacher model.
So, the LGKA module is developed to guide the teacher to build the Visual Semantics Bank (VSB), denoted as $V$, which has the same shape as TSB. For each input, the teacher's visual feature is classified by the TSB and is used to momentum update the VSB. Since the input \textit{[class]} is self-determined, the user can specify the texts containing different semantic categories' knowledge to supervise the teacher to transfer the knowledge to the student.

Lastly, to align the teacher and student in both visual and textual space, the similarity of the outputs of the teacher and student with the feature in TSB and VSB is calculated as the consistency constraint in both appearance and semantic space.

\subsection{Language-Guided Knowledge Aggregation} 
In order to transfer related knowledge from the teacher to the student, the LGKA module is developed to constrain the learning scope.
Specifically, we first treat textual semantics bank $L$ as a classifier and classify the visual feature $z_I$ extracted by the teacher for each batch, which can be formulated as
\begin{equation}
\label{eq1}
    \theta = argmax(z_I \cdot L),
\end{equation}
where $z_I\in \mathbb{R}^{B\times D}$, $L \in \mathbb{R}^{D\times C}$ and $\theta\in \mathbb{R}^{B\times1}$. The $B$, $D$, $C$ and $\theta$ denote batch size, \#channels of the output of the text encoder, the number of self-determined categories, and the classification results of the teacher's visual feature, respectively. 
Note that the categories are customized depending on the user given \textit{[class]}. By calculating the similarity, each image feature will be classified into the most semantically similar category.

There may be more than one feature in a batch $z_I$ that belongs to the same category, then the average vectors are computed, which can be formulated as
\begin{equation}
    z_C^i = mean(z_I[\theta = i]),
\end{equation}
where $z_C\in \mathbb{R}^{D \times C}$ denotes the centroid feature of each semantic category in this batch and $i\in[0, C-1]$. Note that only when one category is contained in the batch, its centroid feature is computed and the visual anchor for this category in VSB is updated. 

At last, the centroid feature of each semantic category $z_C$ will be added to the VSB by momentum updating, which can be formulated as
\begin{equation}
\left\{
    \begin{array}{l}
        V^i \gets m V^i + (1-m)z_C^i, \\
        V^i \gets z_C^i (init.),
    \end{array}
\right.
\end{equation}
where $V \in \mathbb{R}^{D \times C}$. Similar to~\cite{he2020momentum}, $m$ is a momentum coefficient and set to 0.999 as default. When updating the visual anchor of one semantic category in VSB for the first time, we directly replace $V$ with $z_C$ instead of momentum updating it, which can produce a better initial point as compared with updating all classes' features in VSB from randomly initialized points. 

The proposed VSB has two advantages compared with the previous method. 1) Memory-friendly. Previous SSD methods~\cite{abbasi2020compress, fang2021seed, xu2021bag} need to maintain a large queue with enough anchor points to get satisfactory results. Similar to MoCo~\cite{he2020momentum}, the queue length is 65536, containing a large number of similar features for the same semantic category, consuming a lot of memory. The proposed VSB only saves the anchor feature for each semantic category, which is more memory-efficient. 2) Data-efficient. Since previous SSD methods~\cite{abbasi2020compress, fang2021seed} maintain a queue to store features, they can only save features in the neighboring few batches. On the contrary, the proposed VSB saves whole dataset-aware features through momentum updating, showing higher data efficiency.

\subsection{Language-aware Knowledge Transfer}
\label{sec:losses}
As mentioned before, the language information is used to supervise the optimization of the whole model from both visual and textual space, to be detailed below.

\textbf{Visual Space Alignment Loss.}
After LGKA, the VSB maintains the visual anchor for measuring the image-to-image similarity. To align the output of the teacher and the student in visual space, the cosine similarity scores between teacher/student feature $z_T/z_S$ and $V$ are calculated first. And similar to~\cite{fang2021seed}, the teacher feature is added at the end of VSB to directly align the student with the teacher. Therefore, the modified VSB is $V'=[V^0,\cdots,V^{C-1},V^C]$ with $V^C=z_T^i$. The calculation of the similarity score can be formulated as

\begin{align}
    s_{T-V}^i &=\frac{exp(z_T^i \cdot V'/\tau_T)}{\sum_j exp(z_T^i \cdot V'^j/\tau_T)}, \\
    s_{S-V}^i &=\frac{exp(z_S^i \cdot V'/\tau_S)}{\sum_j exp(z_S^i \cdot V'^j/\tau_S)}, 
\end{align}

where $s_{T-V} \in \mathbb{R}^{(C+1)\times B}$ and $s_{S-V} \in \mathbb{R}^{(C+1)\times B}$. And $\tau_T$ and $\tau_S$ is the temperature parameter.

The proposed visual space align loss can be formulated as the cross entropy between $s_{T-V}$ and $s_{S-V}$, which can be formulated as
\begin{align}
    L_{VIS} &= -\sum_{i=0}^{B-1}s_{T-V}^i \cdot log s_{S-V}^i,
\end{align}

\textbf{Textual Space Alignment Loss.}
When language information is introduced, the output of the student should be aligned with the teacher's output not only in visual space but also in textual space. Specifically, 
to this end, this loss directly uses textual anchors in TSB to construct the image-to-language relation, which can be formulated as

\begin{align}
    s_{T-L}^i &=\frac{exp(z_T^i \cdot L/\tau_T)}{\sum_j exp(z_T^i \cdot L^j/\tau_T)}, \\
    s_{S-L}^i &=\frac{exp(z_S^i \cdot L/\tau_S)}{\sum_j exp(z_S^i \cdot L^j/\tau_S)}, \\
    L_{TEX} &= -\sum_{i=0}^{B-1}s_{T-L}^i \cdot log s_{S-L}^i,
\end{align}

where $s_{T-L} \in \mathbb{R}^{C\times B}$ and $s_{S-L} \in \mathbb{R}^{C\times B}$ denote the similarity score between extracted feature $z_T/z_S$ and $z_L$. And $\tau_T$ and $\tau_S$ is the temperature parameter.

In sum, the total loss is denoted as:
\begin{equation}
    L_{LGD} = \alpha L_{VIS} + (1-\alpha) L_{TEX},
\end{equation}
where $\alpha$ is a hyper-parameter to balance the $L_{VIS}$ and $L_{TEX}$, which sets to 0.5 in our default.

\subsection{Generalize to More Image Encoders}
\label{variants}
To verify the generalization of the proposed LGD, we attempt to extend the LGD to other pre-trained image encoders, such as an SSL model, by making corresponding changes to solve two problems.

The first problem is the mismatch of feature dimensions between the image encoder and text encoder. For example, the representation dimension of MoCo~\cite{he2020momentum} and SimSiam~\cite{chen2021exploring} is 128-D, while CLIP is 1024-D. Therefore, we add a learnable MLP layer after the $L$ to downsample the dimension of $L$ from 1024 to 128. The second one is mode collapse. 
From the formula of KL divergence, we can derive that:

\begin{equation}
    L_{TEX} = D_{KL}(s_{T-L}||s_{S-L})+H(s_{T-L}),
\end{equation}

where $D_{KL}$ is the KL-Divergence between similarity scores of T-L and S-L, and $H(s_{T-L})$ denotes the entropy of $s_{T-L}$. When we minimize $L_{TEX}$ as before, since the TSB $L$ is constant and $z_T$ is learnable (due to the added MLP layer to solve the first problem), $z_T$ is easy to fall into the suboptimal solution $\bf{0}$ to make the entropy of $s_{T-L}$ minimal. 
To solve this problem, we let the difference between $T-L$ and $S-L$ mimic the difference between $T-V$.
The total loss can be formulated as
\begin{equation}
    \begin{split}
        L_{LGD} = & \alpha \sum_{i=0}^{B-1}-s_{T-V}^i \cdot log s_{S-V}^i \\
                & +\alpha \sum_{i=0}^{B-1}-s_{T-V}^i \cdot log s_{T-L}^i \\
                & +\alpha \sum_{i=0}^{B-1}-s_{T-V}^i \cdot log s_{S-L}^i,
    \end{split}
\end{equation}
where $\alpha$ is 0.33 in our default.
We use the same $\alpha$ for different parts to keep a balanced contribution from each term in the total loss function.
In this manner, since $z_T$ is learnable, the VSB $V$ is not constant, then $\bf{0}$ will no longer be a suboptimal solution making the loss function falling into a local minima. Additionally, learning with this loss function can minimize the KL-Divergence between similarity scores of T-V and T-L/S-L, thus achieving the original goal of driving the similarity score between $z_S$ and $L$ mimicking that between $z_T$ and $L$. By doing so, the distillation process from teacher to student can be successfully enhanced by the proposed TSB and VSB without the two problems mentioned above.

\subsection{Text Control on Downstream Tasks}
For different downstream tasks and application scenes, the corresponding category names can be utilized as language guidance for distilling processing, rather than the fixed one. With the selected category names and some pre-built templates, the pre-trained language model can provide specific semantic knowledge of each category due to its ability to understand universal text contents. In some open-vocabulary perception works~\cite{qin2023freeseg,liu2023grounding,zhang2022pointclip}, task-specific category names can help refine the image feature and find the patterns with corresponding semantics. In our setting, taking the scene in Section~\ref{sec:losses} as an example, minimizing $L_{TEX}$ is equivalent to minimizing $D_{KL}(s_{T-L}||s_{S-L})$. If $L$ is built upon category names that not match the image data distribution of current dataset, then values of $s_{T-L}$ in those mismatched dimensions will always have low values, preventing student networks from learning useful information from distribution alignment. When using the selected task-oriented category names, ideal knowledge transfer can be achieved for each category in current scenario.

%------------------------------------------------------------------------
\section{Experiment}
\label{4_exp}
\subsection{Setup}

\textbf{Dataset and Downstream Tasks.}
For model pre-training with the LGD, the ImageNet-1k (IN-1k)~\cite{deng2009imagenet} dataset is used as unlabeled source data by abandoning the labels, which contains 1.28M images for training, and 50,000 images for validation. Then, we evaluate the pre-trained model on various downstream tasks, including zero-shot and fully-supervised classification on the IN-1k and the Caltech-256 (CT-256)~\cite{griffin_holub_perona_2022} dataset, semi-supervised classification on the IN-1k dataset, object detection and instance segmentation on the COCO 2017~\cite{lin2014microsoft} dataset, long-tailed object detection and instance segmentation on the LVIS v1~\cite{gupta2019lvis} dataset, and semantic segmentation on the ADE20K\cite{zhou2017scene} and Cityscapes (CS)~\cite{cordts2016cityscapes} datasets.

\textbf{Teacher-Student Pair.}
Experiments are mainly conducted on two pairs of teacher-student models: CLIP pre-trained ResNet-50 (CLIP RN50)~\cite{radford2021learning} → ResNet-18 (RN18)~\cite{he2016deep} and CLIP RN50~\cite{radford2021learning} → MobileNetV2 (MNV2)~\cite{sandler2018mobilenetv2}, representing knowledge transfer between similar and dissimilar networks respectively. Besides, we also conduct experiments with different teacher networks, such as MoCo RN50~\cite{he2020momentum} and SimSiam RN50~\cite{chen2021exploring}. The pre-trained text encoder used is a Transformer~\cite{vaswani2017attention} based encoder that is jointly trained with CLIP RN50.

\begin{table*}[!t]
  \caption{Semi/Fully-supervised classification results on the ImageNet-1k (IN-1k) and Caltech-256 (CT-256) datasets. The top-1 accuracy is reported.}
  \begin{center}
  \begin{tabular}{c|ccccc|cc|cccc}
    \toprule
    \multirow{3}{*}{\textbf{Method}} & \multirow{3}{*}{\textbf{Teacher}} & \multirow{3}{*}{\textbf{Student}} & \multirow{3}{*}{\textbf{Data}} & \multirow{3}{*}{\textbf{Text}} & \multirow{3}{*}{\textbf{Epoch}} & \multicolumn{2}{c|}{\textbf{Zero-shot}} & \multicolumn{4}{c}{\textbf{Linear-probe}} \\
     & & & & & & \multirow{2}{*}{\textbf{IN-1k}} & \multirow{2}{*}{\textbf{CT-256}} & \multicolumn{3}{c}{\textbf{IN-1k}} & \multirow{2}{*}{\textbf{CT-256}} \\
     & & & & & & & & 1\% & 10\% & 100\% & \\
    \midrule
    MoCo & RN50 & - & - & - & - & - & - & - & - & 67.5 & - \\
    CLIP & RN50 & - & - & - & - & 58.9 & 78.1 & - & - & 73.3 & - \\
    \midrule
    Super. & - & RN18 & IN-1k & - & 90 & - & - & -& - & - & 77.1\\
    SEED & MoCo RN50 & RN18 & IN-1k & - & 200 & - & - & 37.5 & 51.1 & 57.9 & 78.5\\
    BINGO & MoCo RN50 & RN18 & IN-1k & - & 200 & - & - & 42.8 & 57.5 & 61.4 & 79.3 \\
    % KDEP & MS RN50 & RN18 & IN-1k & - & 90 & 39.9 & 52.3 & 58.8 & 79.1 \\
    DisCo & MoCo RN50 & RN18 & IN-1k & - & 200 & - & - & - & - & 60.6 & - \\
    SMD & SimSiam RN50 & RN18 & IN-1k & - & 100 & - & - & - & - & 61.8 & - \\
    SEED & CLIP RN50 & RN18 & IN-1k & - & 200 & 44.2 & 62.1 & 43.1 & 56.3 & 62.7 & 79.5\\
    BINGO & CLIP RN50 & RN18 & IN-1k & - & 200 & 45.5 & 62.3 & 44.2 & 59.2 & 64.0 & 80.6\\
    \textbf{LGD} & CLIP RN50 & RN18 & IN-1k & IN-1k & 200 & \textbf{49.5} & 63.9 & \textbf{53.5} & \textbf{61.7} & \textbf{67.2} & \textbf{82.7} \\
    \textbf{LGD} & CLIP RN50 & RN18 & IN-1k & IN-1k & 90 & 47.8 & 62.3 & \textbf{52.9} & \textbf{61.5} & \textbf{66.4} & \textbf{81.6} \\
    
    \textbf{LGD} & CLIP RN50 & RN18 & IN-1k & CT-256 & 90 & 34.3 & 65.8 & - & - & - & - \\
    \textbf{LGD} & CLIP RN50 & RN18 & CT-256 & CT-256 & 90 & 18.9 & \textbf{66.5} & - & - & - & - \\
    \midrule
    Super. & - & MNv2 & IN-1k & - & 90 & - & - & - & - & - & 77.5\\
    SEED & CLIP RN50 & MNv2 & IN-1k & - & 200 & - & - & 42.2 & 55.5 & 63.5 & 79.1\\
    BINGO & CLIP RN50 & MNv2 & IN-1k & - & 200 & - & - & 43.5 & 56.9 & 64.6 & 80.1\\
    \textbf{LGD} & CLIP RN50 & MNv2 & IN-1k & IN-1k & 200 & \textbf{50.3} & 64.7 & \textbf{51.9} & \textbf{60.7} &\textbf{67.4} & \textbf{81.3} \\
    \textbf{LGD} & CLIP RN50 & MNv2 & IN-1k & IN-1k & 90 & 48.5 & 63.0 & \textbf{50.3} & \textbf{59.4} & \textbf{66.3} & \textbf{81.2}\\
    \textbf{LGD} & CLIP RN50 & MNv2 & IN-1k & CT-256 & 90 & 35.0 & 66.8 &-&-&-& -\\
    \textbf{LGD} & CLIP RN50 & MNv2 & CT-256 & CT-256 & 90 & 19.2 & \textbf{67.4} & - &-&-&- \\
    \bottomrule
  \end{tabular}
  \end{center}
  \label{main2}
  % \vspace{-5mm}
\end{table*}
\begin{table*}[!t]
  \caption{The results of object detection and instance segmentation on the COCO and LVIS datasets, and the results of semantic segmentation on the ADE20K and Cityscapes (CS) datasets.}
  \begin{center}
  \begin{tabular}{c|ccccc|cccccc}
    \toprule
    \multirow{2}{*}{\textbf{Method}} & \multirow{2}{*}{\textbf{Teacher}} & \multirow{2}{*}{\textbf{Student}} & \multirow{2}{*}{\textbf{Data}} & \multirow{2}{*}{\textbf{Text}} & \multirow{2}{*}{\textbf{Epoch}} & \multicolumn{2}{c}{\textbf{COCO}} & \multicolumn{2}{c}{\textbf{LVIS}} & \textbf{ADE20K} & \textbf{CS}\\
     & & & & & & $AP^{bb}$ & $AP^{mk} $ & $AP^{bb}$ & $AP^{mk}$ & $mIoU$ & $mIoU$\\
    \midrule
    MoCo & RN50 & - & IN-1k & - & - & 38.5 & 35.1 & 22.7 & 21.9 & 38.9 & 75.3 \\
    CLIP & RN50 & - & WIT & - & - & 39.3 & 36.8 & 23.0 & 22.1 & 39.6 & 75.8 \\
    \midrule
    Super. & - & RN18 & IN-1k & - & 90 & 32.8 & 31.3 & 17.5 & 18.6 & 33.3 & 71.3 \\
    SEED & MoCo RN50 & RN18 & IN-1k & - & 200 & 33.7 & 31.8 & 18.1 & 19.2 & 34.0 & 72.2\\
    BINGO & MoCo RN50 & RN18 & IN-1k & - & 200 & 33.9 & 32.1 & 18.4 & 19.3 & 34.4 & 72.3\\
    SEED & CLIP RN50 & RN18 & IN-1k & - & 200 & 34.4 & 32.2 & 18.9 & 19.7 & 34.4 & 72.8\\
    BINGO & CLIP RN50 & RN18 & IN-1k & - & 200 & 34.6 & 32.4 & 19.2 & 19.9 & 34.7 & 73.2 \\
    \textbf{LGD} & CLIP RN50 & RN18 & IN-1k & IN-1k & 90 & \textbf{34.0} & \textbf{32.0} & \textbf{19.7} & \textbf{20.3} & \textbf{34.9} & \textbf{73.7} \\
    \textbf{LGD} & CLIP RN50 & RN18 & IN-1k & IN-1k & 200 & \textbf{35.1} & \textbf{32.9} & \textbf{20.1} & \textbf{20.5} & \textbf{35.3} &\textbf{74.0}\\
    \midrule
    Super. & - & MNv2 & IN-1k & - & 90 & 27.3 & 26.4 & 14.9 & 14.4 & 29.7 & 70.2 \\
    SEED & MoCo RN50 & MNv2 & IN-1k & - & 200 & 28.1 & 26.8 & 15.4 & 14.8 & 31.9 & 70.5\\
    BINGO & MoCo RN50 & MNv2 & IN-1k & - & 200 & 28.3 & 27.0 & 15.6 & 15.0 & 32.2 & 70.8\\
    % KDEP & MS RN50 & MNv2 & IN-1M & - & 90 & 28.3 & 27.1 & 15.7 & 15.8 & 32.3 & 71.0 \\
    SEED & CLIP RN50 & MNv2 & IN-1k & - & 200 & 28.9 & 27.4 & 15.9 & 16.1 & 32.2 & 71.0 \\
    BINGO & CLIP RN50 & MNv2 & IN-1k & - & 200 & 29.1 & 27.7 & 16.0 & 16.2 & 32.5 & 71.3 \\
    \textbf{LGD} & CLIP RN50 & MNv2 & IN-1k & IN-1k & 90 & \textbf{28.8} & \textbf{27.5} & \textbf{16.1} &\textbf{16.6} & \textbf{33.8} & \textbf{72.3}  \\
    \textbf{LGD} & CLIP RN50 & MNv2 & IN-1k & IN-1k & 200 & \textbf{29.6} & \textbf{28.1} & \textbf{16.2} & \textbf{16.9} & \textbf{33.9} & \textbf{72.3}\\
    \bottomrule
  \end{tabular}
  \vspace{0mm}
  \end{center}
  \label{main3}
  % \vspace{-5mm}
\end{table*}

\textbf{Comparison Methods.}
The proposed LGD is compared with several small model pre-training methods: 1) fully-supervised pre-training method, including ImageNet pre-training~\cite{he2019rethinking}; 2) SSD based methods, including SEED~\cite{fang2021seed}, BINGO~\cite{xu2021bag}, DisCo~\cite{gao2022disco}, and SMD~\cite{liu2022improving}.

\textbf{Implementation Details.} \label{implementation}
The proposed LGD is implemented using the PyTorch framework, and all experiments are conducted on 8 NVIDIA TESLA V100 GPUs. The distillation process is trained with a standard SGD optimizer with a momentum of 0.9 and a weight decay parameter of 1e-4 for 90 and 200 epochs.
The initial learning rate is set as 0.03 and updated by a cosine decay scheduler~\cite{nair2010rectified} with 5 warm-up epochs and a batch size of 256 per GPU. 

For transferring to image classification, we conduct the supervised linear classification on IN-1k and CT-256. For zero-shot setting, we treat the textual semantics bank $L$ as a classifier and the predicted category index is calculated following Eq.~\ref{eq1}. In this way, each image feature will be classified into the most semantically similar category without training with the corresponding label. For linear-probe setting, following previous works~\cite{he2020momentum, fang2021seed}, we train a single linear layer classifier on top of the frozen network encoder after distillation or pre-training. For IN-1k dataset, besides fully supervised learning which utilizes 100\% of the training set, following previous works~\cite{chen2020simple,xu2021bag}, we evaluate the proposed method by fine-tuning the student model with 1\% and 10\% labeled data as semi-supervised setting. We follow the training split settings as in previous works for fair comparisons. SGD optimizer is used to prepare the linear classifier for 100 epochs with a weight decay of 0. The initial learning rate is set as 30 and is then reduced by a factor of 10 at 60 and 80 epochs. The results are reported in terms of Top-1 accuracy.

For transferring to object detection and instance segmentation, we use mmdetection~\cite{mmdetection} for implementation. We train Mask-RCNN FPN~\cite{he2017mask} with RN18 and MNv2 backbone to evaluate the transferability of the learned features on COCO 2017 and LVIS v1. AdamW~\cite{loshchilov2017decoupled} optimizer is used to finetune the whole network for 12 epochs (the default 1x schedule). The initial learning rate is set as 2e-4 and then reduced by a factor of 10 at 8 and 11 epochs. To better preserve the pre-trained weights, we set the learning rate of the image encoder as 1/10 of the other parameters.

For semantic segmentation, we use mmsegmentation~\cite{mmsegmentation} for implementation. We train Semantic FPN~\cite{kirillov2019panoptic} with RN18 backbone and PSPNet\cite{zhao2017pyramid} with MNv2 backbone to evaluate the transferability of learned features on ADE20k and CS. AdamW~\cite{loshchilov2017decoupled} optimizer is also used, and we also set the learning rate of the image encoder as 1/10 of the other parameters. For MNv2 and RN18 backbone, the initial learning is set as 1e-2 and 1e-4, respectively. For ADE20k and Cityscapes, we train the model for 160k and 80k iterations, respectively.

\subsection{Main Results}

\subsubsection{Results on Zero-shot Classification.}
% \subsubsection{4.2.1 Implementation Details}
% \subsubsection{4.2.2 Results}
As shown in Table~\ref{main2}, we compare our method with previous SSD methods SEED~\cite{fang2021seed} and BINGO~\cite{xu2021bag} on the ImageNet-1k and Caltech-256 validation set. The top-1 zero-shot classification accuracy is reported. We observe that the proposed LGD surpasses the contrastive SSD methods consistently on all benchmarks, verifying the effectiveness of the proposed LGD.

To further verify the effect of language guidance, we still use the same IN-1k as unlabeled image input but change the input texts to the class of CT-256 for doing experiments. Significant improvement is observed on both RN18 (+3.5\%) and MNv2 (+3.8\%) backbone, which shows that we can specify the VSB and TSB through the texts to control the knowledge we want to transfer to the students. Besides, we use the CT-256 as our unlabeled image input and achieve the highest zero-shot accuracy (66.5\% and 67.4\%) on the CT-256 dataset, which reveals that the consistency of data between the distillation and downstream tasks will impact the results. More relevance between the source data used in distillation and the downstream task brings better downstream performance of the trained small model. Nevertheless, to fairly compare the performance of small models transferring to downstream tasks with other methods, we uniformly use IN-1k as the source data for distillation or pre-training in the following experiments.

\begin{figure}[t]
\centering
\includegraphics[width=\columnwidth]{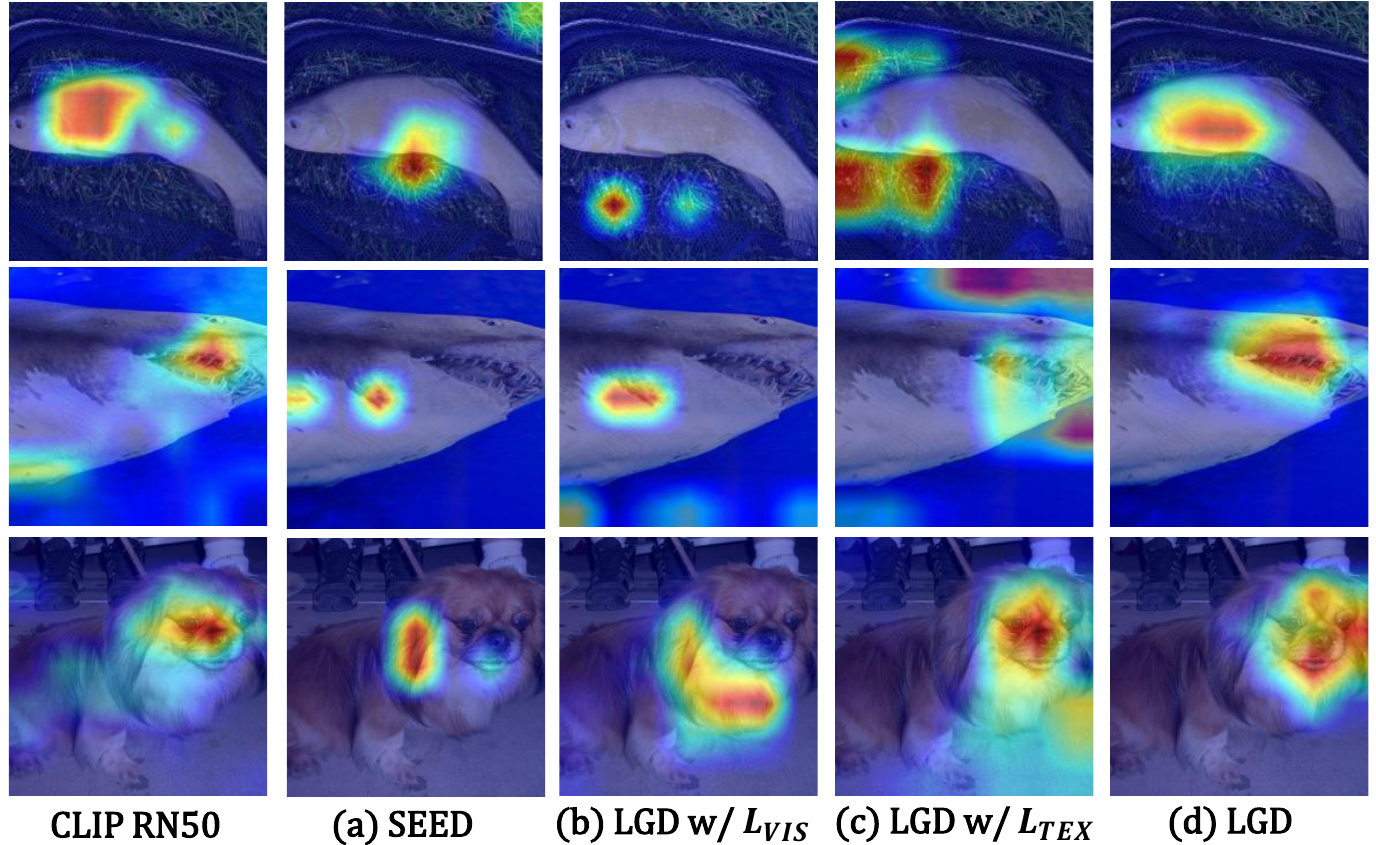} 
\caption{The Grad-CAM visualization shows the different attention maps of distilled RN18 in columns (a)-(d). The first column shows the attention map of CLIP pre-trained RN50, which is the teacher model. Columns (a)-(d) show the visualization results of SEED, LGD with only visual space alignment, LGD with only textual space alignment, and LGD with both losses, respectively.}
\label{fig3}
% \vspace{-3mm}
\end{figure}

\subsubsection{Transfer to Semi/Fully-supervised Classification.} \label{exp-classification}
Following~\cite{fang2021seed, xu2021bag}, we evaluate the learned representation on semi-supervised/fully-supervised classification on IN-1k, where a fixed 1\%, 10\% or 100\% of IN-1k training data are provided with the annotations. Besides, to further study whether the improvement of the learned representations by distillation is confined to ImageNet, we evaluate the additional classification dataset CT-256 to study the generalization and transferability of the feature representation. As shown in Table~\ref{main2}, the proposed LGD has a remarkably 1.2\%-9.3\% improvement compared with previous methods.

\subsubsection{Transfer to Object Detection and Instance Segmentation.}
As shown in Table~\ref{main3}, we conduct experiments on two downstream tasks including object detection and instance segmentation, on the COCO 2017 and LVIS v1 datasets. Compared with standard ImageNet supervised pre-training (Super.), the distilled model pre-trained by the proposed LGD achieves a large improvement in the same number of training epochs. On COCO, the RN18 based Mask RCNN shows +1.2 and +0.7 point improvement on $AP^{bb}$ and $AP^{mk}$, respectively. And the MNv2 based Mask RCNN shows +1.5 and +1.7 point improvement on $AP^{bb}$ and $AP^{mk}$. Compared with other SSD methods, our method also shows a significant improvement. Compared with SEED and BINGO pre-trained RN18, which is distilled from CLIP RN50, the proposed LGD shows a consistent +0.5 to +0.9 point improvement on COCO. Compared with the RN18 distilled from MoCo RN50, the proposed LGD shows a large improvement of +0.8 to +1.7 points. The MNV2 backbone also shows a similar improvement and the detailed experiment results can be seen in Table~\ref{main3}.

\subsubsection{Transfer to Semantic Segmentation.}
As shown in Table~\ref{main3}, we conduct semantic segmentation experiments on the ADE20k and Cityscapes (CS) datasets. Compared with standard ImageNet supervised pre-training (Super.), the RN18 based Semantic FPN shows +1.6 and +2.4 mIoU improvement on the two datasets, respectively. Besides, compared with SSD methods, the proposed LGD also shows significant +0.5 to +1.8 mIoU improvement. For the MNv2 backbone, it also shows a similar improvement and the detailed experiment results can be seen in Table~\ref{main3}. 

Notably, we observe that the proposed LGD surpasses the comparative methods, including standard ImageNet pre-training and previous self-supervised distillation, on all benchmarks with the same overhead of training time and data amount. This also proves the generalization ability of the learned representations from language-guided distillation to a wide range of data domains and classes.

\subsection{Ablation and Analysis}

\subsubsection{Different Distillation Strategies.}

\begin{table}[!t]
  \caption{Zero-shot IN-1k Top-1 Acc.(\%) of the distilled RN18 with different distillation strategies.}
  \begin{center}
  \begin{tabular}{c|c|c}
    \toprule
    \textbf{Method} & \textbf{Loss} & \textbf{ZS Top-1} \\
    \midrule
    SEED & - & 44.2 \\
    \textbf{LGD} & $L_{VIS}$ & 45.7 \\
    \textbf{LGD} & $L_{TEX}$ & 47.3 \\
    \textbf{LGD} & $L_{VIS}+L_{TEX}$ & 47.8 \\
    \bottomrule
    % \hline
  \end{tabular}
  \vspace{0mm}
  \end{center}
  \label{ablation1}
  % \vspace{-3mm}
\end{table}
\begin{table}[!t]
  \caption{The results of evaluating the text control on other downstream tasks, including linear-probe top-1 acc.(\%) on the Caltech-256 (CT-256), and the mIoU on the Cityscapes (CS) and ADE20K (ADE).}
  \begin{center}
  \begin{tabular}{c|c|c|c|c|c}
    \toprule
    \textbf{Method} & \textbf{Data} & \textbf{Text} & \textbf{CT-256} & \textbf{CS} & \textbf{ADE}\\
    \midrule
    LGD & IN-1k & IN-1k & 81.6 & 73.7 & 34.9\\
    LGD & IN-1k & CT-256 & 81.9 & - & -\\
    LGD & IN-1k & CS & - & 74.1 & -\\
    LGD & IN-1k & ADE & - & - & 35.2\\
    \bottomrule
  \end{tabular}
  \vspace{0mm}
  \end{center}
  \label{ablation2}
  % \vspace{-5mm}
\end{table}

To show the effectiveness of consistency loss in visual and textual space, we conduct an ablation study and show results in Table~\ref{ablation1}. We use CLIP RN50 as the teacher network and report the zero-shot top-1 accuracy on the IN-1k validation set. SEED~\cite{fang2021seed} trains a student to mimic the similarity score distribution inferred by a teacher over a set of randomly maintained instances. $L_{VIS}$ and $L_{TEX}$ are the proposed consistency losses in visual and textual space. Compared with SEED, the proposed visual constraint $L_{VIS}$ has a significant +1.5\% improvement. With the constraint in both visual and textual space ($L_{VIS}$ + $L_{TEX}$), we can get the best result of 47.8\%.

To further explore the effect of different distillation strategies, the attention maps are visualized by Grad-CAM~\cite{selvaraju2017grad} and shown in Fig.~\ref{fig3}. The proposed LGD has the most accurate attention maps and the most similar attention area to the teacher (CLIP RN50). And we find that if we only apply constraints in visual space, such as SEED and $L_{VIS}$, the student network will pay attention to the general feature of the category. For example, it will pay attention to the net and sea when it identifies the fish and shark. Although images of different categories commonly share these areas, they can not sufficiently represent the semantic categories in the downstream task. Besides, with the constraint in both visual and textual space, the attention map can focus on the most critical area, such as the animals' faces.

\subsubsection{Text Control on More Downstream Tasks} \label{exp-downstream}

\begin{table}[t]
  \caption{Results of different architecture for teacher and student on IN-1k Top-1 Acc.(\%).}
  \begin{center}
  % \small
  \begin{tabular}{c|c|c|c}
    \toprule
    \textbf{Method} & \textbf{Teacher} & \textbf{Student} & \textbf{Top-1} \\
    \midrule
    
    CLIP & - & ViT-B/16 & 68.3 \\
    SEED & CLIP RN101 & RN18 & 45.2 \\
    \textbf{LGD} & CLIP RN101 & RN18 & 50.3 \\
    SEED & CLIP ViT-B/16 & RN18 & 53.2 \\
    \textbf{LGD} & CLIP ViT-B/16 & RN18 & 55.9 \\
    SEED & CLIP ViT-B/16 & DeiT-tiny & 53.3 \\
    \textbf{LGD} & CLIP ViT-B/16 & DeiT-tiny &  56.5\\
    \bottomrule
  \end{tabular}
  \vspace{0mm}
  \end{center}
  \label{ablation4}
\end{table}
\vspace{0mm}
\begin{table}[!t]
   \caption{Linear-probe IN-1k Top-1 Acc.(\%) of distilled RN18. LGD is combined with other SSL pre-trained image and text encoder.}
  \begin{center}
  \begin{tabular}{c|c|c|c}
    \toprule
    \textbf{Method} & \textbf{Image Encoder} & \textbf{Text Encoder} & \textbf{Top-1} \\
    \midrule
    SEED & MoCo RN50 & - & 57.9 \\
    \zl{SEED+\textbf{LGD}} & MoCo RN50 & CLIP text & 59.7 \\
    \zl{SEED+\textbf{LGD}} & MoCo RN50 & $BERT_{BASE}$ & 58.5\\
    SMD & SimSiam RN50 & - & 61.8 \\
    \zl{SMD+\textbf{LGD}} & SimSiam RN50 & CLIP text &  62.1\\
    \bottomrule
  \end{tabular}
  \vspace{0mm}
  \end{center}
  \label{ablation3}
  %\vspace{-5mm}
\end{table}

In this section, we evaluate the text control on more downstream tasks as shown in Table~\ref{ablation2}. We firstly distill a small model and then finetune the distilled small model on the downstream tasks as described in Section~\ref{implementation}. In the distillation process, we use the IN-1k as unlabeled source data and the language prompt with class names in the downstream tasks as text input. For example, there are 19 text inputs for Cityscapes, 150 text inputs for ADE20K, and 256 text inputs for Caltech-256. The CLIP RN50 and RN18 are selected as the teacher and student models. And the training epoch is set to 90. Other hyper-parameters are the same as those introduced in Section~\ref{implementation}. 

Constant improvements are observed on the Caltech-256 (+0.3), Cityscapes (+0.4), and ADE20K (+0.3). It shows that through the text control, the small model can learn more useful knowledge from the teacher and perform better on corresponding downstream tasks.

\subsubsection{Different Pre-trained Image/Text Encoder.}

To evaluate the generalization of the proposed LGD, we combine LGD with other SSL pre-trained image or text encoder and show consistent improvement. Specifically, we introduce LGD into SEED~\cite{fang2021seed} and SMD~\cite{liu2022improving} as the language supervision module introduced in the above Section~\ref{variants}. As shown in Table~\ref{ablation3}, SEED + LGD has a significant +1.8\% improvement compared with SEED. And SMD + LGD also has a +0.3\% improvement compared with SMD. 
Besides, we use BERT~\cite{jacob2019bert} as text encoder, which also shows +0.6\% improvement compared with SEED.
Note that in the above experiments, the image encoder and text encoder are self-supervised and pre-trained independently, which shows that the proposed LGD also works on non-joint-trained image and text encoder.

\begin{table}[t]
  \caption{Training and Testing Time Comparison between LGD and SEED.}
  \begin{center}
  % \small
  \begin{tabular}{c|c|c|c}
    \toprule
    \textbf{Method} & \textbf{Pre-training (h)} & \textbf{Finetuning (h)} & \textbf{Testing (m)} \\
    \midrule
    
    SEED & 58.8(200) & 14.8 & 2.5 \\
    LGD & 36.3(90) / 72.2(200) & 14.7 & 2.7 \\

    \bottomrule
    % \hline
  \end{tabular}
  \vspace{0mm}
  \end{center}
  \label{ablation_time}
\end{table}
\vspace{0mm}

\begin{table}[t]
  \caption{Zero-shot and Linear-probe classification results on ModelNet40 dataset. The overall accuracy(\%) is reported.}
  \begin{center}
  % \small
  \begin{tabular}{c|c|c|c}
    \toprule
    \textbf{Method} & \textbf{Student} & \textbf{Zero-shot} & \textbf{Linear-probe} \\
    \midrule
    
    SEED & PointMLPElite & 78.48 & 89.90 \\
    SEED & Transformer-6L & 77.31 & 89.74 \\
    \textbf{LGD} & PointMLPElite & \bf{80.35} & 91.17 \\
    \textbf{LGD} & Transformer-6L & 79.86 & \bf{91.26} \\

    \bottomrule
  \end{tabular}
  \vspace{0mm}
  \end{center}
  \label{experiment_3d}
\end{table}
\vspace{0mm}

To further verify the applicability of our proposed method across various architectures, we conduct experiments employing CLIP RN101 or CLIP ViT-B/16 as teacher and RN18 or DeiT-tiny as student. As shown in Table~\ref{ablation4}, our proposed LGD demonstrates remarkable performance compared to SEED~\cite{fang2021seed} when applied to multiple teacher/student architectures. For instance, in the cases where we use CLIP ViT-B/16 as teacher and DeiT-tiny as student, LGD achieves notable improvements of 3.2\% compared with SEED. The results of zero-shot ImageNet classification demonstrate that LGD can not only perform well on CNN-based architectures but also show good performance on other architectures, such as ViTs.

\subsubsection{Training and testing time analysis}

The training and testing time of our proposed LGD and previous method SEED~\cite{fang2021seed} is assessed in Table~\ref{ablation_time}. Specifically, both models are trained for 200 epochs and finetuned for 100 epochs on the ImageNet1K dataset under the same settings (batch size, number of GPUs, etc). For testing time, we measure the inference time required for classifying images in the validation dataset.
Results indicate that under the same number of training epochs (200), the pre-training time for LGD is longer than that for SEED. However, it is noteworthy that our method achieves comparable results with just 90 epochs of training (taking \textbf{36.3} hours), as opposed to SEED, which requires 200 epochs (taking \textbf{58.8} hours), as shown in Table~\ref{main2}. Therefore, despite longer pre-training time per epoch, the overall efficiency of LGD is superior, requiring significantly less training time to achieve comparable performance.
Additionally, the finetuning and testing time of SEED and LGD is similar, as both methods only utilize the student network.

\subsubsection{Results on Tasks other than Image}

To demonstrate the efficacy and generalization ability of the proposed LGD, extensive experiments are conducted on point cloud classification task. For teacher network, we select the latest point-language pre-training framework Point-Bind~\cite{guo2023point} with I2P-MAE~\cite{zhang2023learning} as the point cloud encoder. For student, a solid lightweight point cloud encoder PointMLPElite~\cite{ma2022rethinking} is employed. Besides, plain transformer network used in~\cite{yu2022point} with less (6) layers is also used as student since it has a similar structure with teacher point cloud encoder. Results are shown in Table~\ref{experiment_3d}. Compared to the previous method SEED, LGD performs better on both zero-shot and linear-probe classification on ModelNet40 dataset. And the improvement is consistent among both students.

\subsubsection{Visualization of Feature Distributions}

\begin{figure}[t]
\centering
\includegraphics[width=\columnwidth]{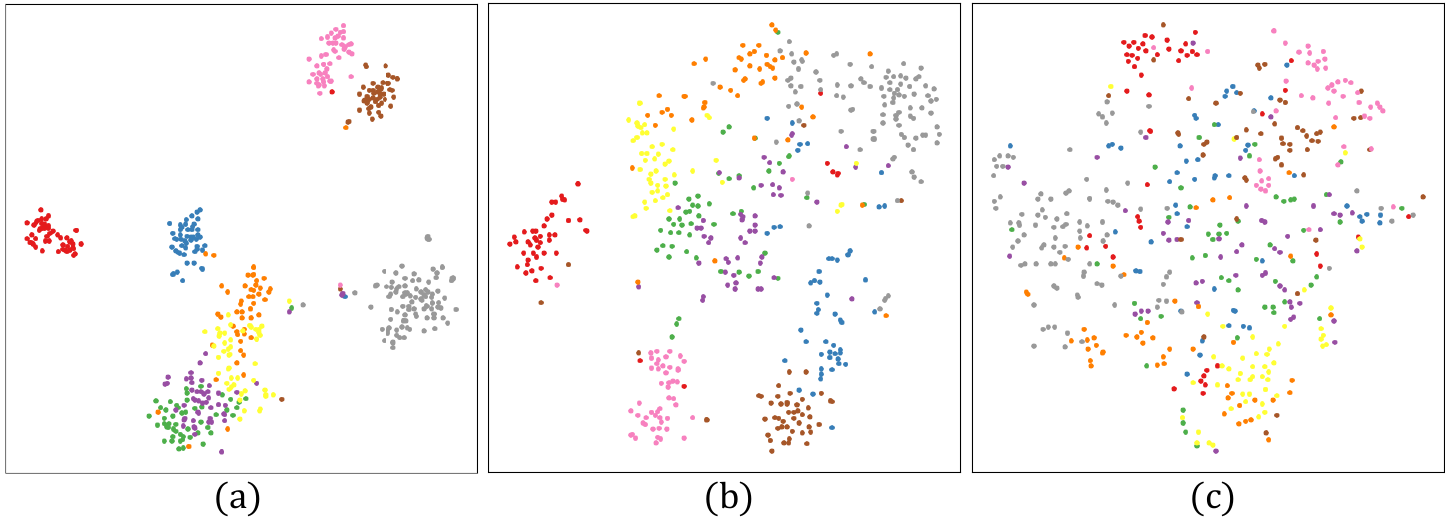} % Reduce the figure size so that it is slightly narrower than the column. Don't use precise values for figure width.This setup will avoid overfull boxes.
\caption{Visualization of feature distributions. (a) CLIP RN50 (teacher); (b) LGD RN18; (c) SEED RN18. Different colors represent data samples of different classes.
Zoom in for details. Best viewed in color.}
\label{fig1}
%\vspace{-2mm}
\end{figure}

\begin{figure}[t]
\centering
\includegraphics[width=\columnwidth]{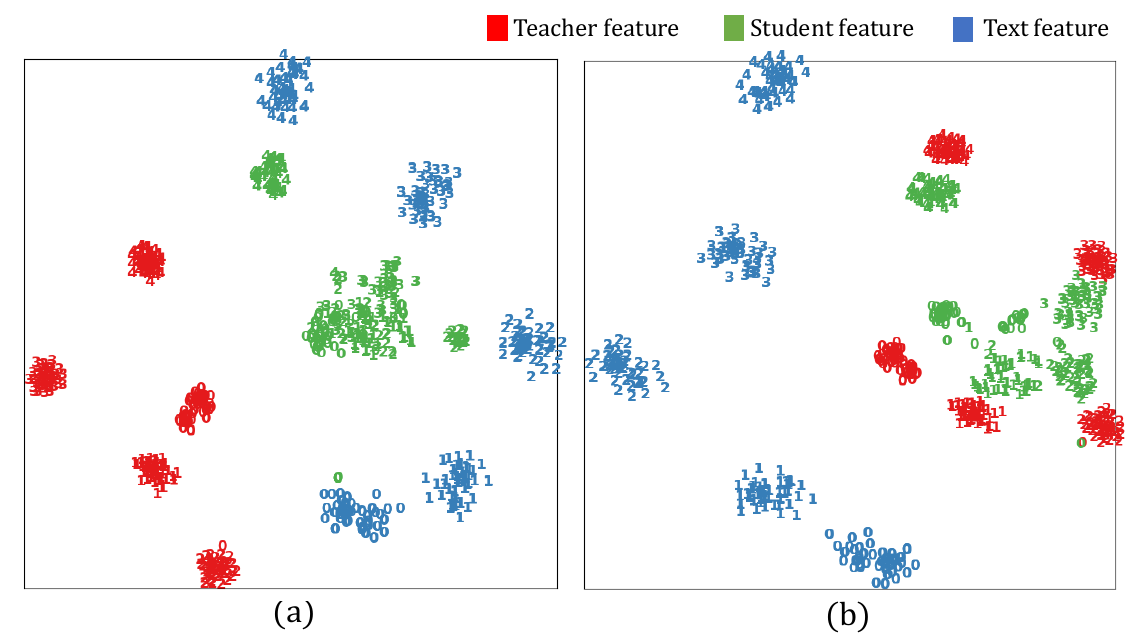} % Reduce the figure size so that it is slightly narrower than the column. Don't use precise values for figure width.This setup will avoid overfull boxes.
\caption{Visualization of feature distributions. (a) LGD; (b) SEED. Different colors represent different types of features, including teacher feature, student feature and text feature. The numbers represent the classes of data samples. Zoom in for details. Best viewed in color.}
\label{fig2}
%\vspace{-2mm}
\end{figure}

To validate the effectiveness of LGD in aligning the outputs of the teacher and student in both the visual and textual semantic spaces, we visualize the feature distribution of the teacher model (CLIP RN50) and student model (LGD RN18 \& SEED RN18~\cite{fang2021seed}).  In Fig.~\ref{fig1}, it can be seen that the LGD learns more compact feature distribution than SEED. Due to the model capacity, the feature distributions of both LGD RN18 and SEED RN18 are more diverse than that of CLIP RN50. In Fig.~\ref{fig2}, the feature distribution of the image encoder (teacher), text encoder, and student are shown. It can be observed that the feature distribution of SEED RN18 is similar to that of the teacher but has a large gap with the text feature of the text encoder which contains task-related knowledge . As shown in Fig.~\ref{fig2} (a), with the constraint in textual space, the feature distributions of LGD RN18 have some variances to that of the teacher but also have a matching relationship with the text feature, representing the combination of knowledge of both visual and textual features.

\section{Conclusion}
\label{5_con}
This paper proposes Language-Guided Distillation (LGD) for transferring knowledge from a pre-trained large teacher model, such as CLIP, to a small one. We first use language guidance to determine which knowledge of the teacher should be transferred to the students. Then, we propose a language-guided knowledge aggregation module to maintain language-guided textual and visual semantic banks. At last, we design two distillation losses to maintain the consistency of teacher and student in both visual and textual space. 
Thorough experiments show that LGD offers significant performance improvement on various downstream tasks.

In our study, we find that the LGD performance largely relies on the joint-trained text and image encoder, while the performance gain on the non-joint-trained one is relatively small because of the mismatch between the textual and visual features.  
We hope our work could promote the development of distillation methods to become a new paradigm for small model pre-training by making full use of the off-the-shelf pre-trained models.

\section*{ACKNOWLEDGMENTS}

This work is supported by National Key R\&D Program of China (2022ZD0160300), National Natural Science Foundation of China (No. 62101137 and 62071127),  Shanghai Natural Science Foundation (No. 23ZR1402900). The computations in this research were performed using the CFFF platform of Fudan University.

\bibliographystyle{IEEEtran}
\bibliography{arxiv}

\vfill

\end{document}